\documentclass[conference]{IEEEtran}
\IEEEoverridecommandlockouts

\usepackage{subcaption}

\usepackage{cite}
\usepackage{amsmath,amssymb,amsfonts,amsthm}

\usepackage{algorithmic}
\usepackage{graphicx}
\usepackage{textcomp}
\usepackage{xcolor}
\usepackage{pgfplots}
\usepackage{float}
\pgfplotsset{compat=1.18}
\def\BibTeX{{\rm B\kern-.05em{\sc i\kern-.025em b}\kern-.08em
    T\kern-.1667em\lower.7ex\hbox{E}\kern-.125emX}}
\begin{document}

\title{Orthogonal Activation with Implicit Group-Aware Bias Learning for Class Imbalance}
\author{\IEEEauthorblockN{1\textsuperscript{st} Sukumar Kishanthan}
\IEEEauthorblockA{\textit{Dialog Axiata PLC} \\
Colombo, Sri Lanka \\
kishanthansukumar@gmail.com}
\and
\IEEEauthorblockN{2\textsuperscript{nd} Asela Hevapathige}
\IEEEauthorblockA{\textit{School of Computing} \\
\textit{Australian National University}\\
Canberra, Australia \\
asela.hevapathige@anu.edu.au}
}

\maketitle

\begin{abstract}
Class imbalance is a common challenge in machine learning and data mining, often leading to suboptimal performance in classifiers. While deep learning excels in feature extraction, its performance still deteriorates under imbalanced data. In this work, we propose a novel activation function, named OGAB, designed to alleviate class imbalance in deep learning classifiers. OGAB incorporates orthogonality and group-aware bias learning to enhance feature distinguishability in imbalanced scenarios without explicitly requiring label information. Our key insight is that activation functions can be used to introduce strong inductive biases that can address complex data challenges beyond traditional non-linearity. Our work demonstrates that orthogonal transformations can preserve information about minority classes by maintaining feature independence, thereby preventing the dominance of majority classes in the embedding space. Further, the proposed group-aware bias mechanism automatically identifies data clusters and adjusts embeddings to enhance class separability without the need for explicit supervision. Unlike existing approaches that address class imbalance through preprocessing data modifications or post-processing corrections, our proposed approach tackles class imbalance during the training phase at the embedding learning level, enabling direct integration with the learning process. We demonstrate the effectiveness of our solution on both real-world and synthetic imbalanced datasets, showing consistent performance improvements over both traditional and learnable activation functions.
\end{abstract}

\begin{IEEEkeywords}
activation functions, class imbalance, deep learning, group-aware bias learning, orthogonality
\end{IEEEkeywords}

\section{Introduction}

Class imbalance is a fundamental challenge in machine learning and data science, as a skewed class distribution often causes classifiers to favor majority classes \cite{krawczyk2016learning}. This bias leads to unreliable predictions, making the models less effective for real-world applications, where class imbalance is a common issue in many real-world datasets.  While deep learning has demonstrated remarkable success in feature extraction and generalizability across various downstream tasks, its performance has also been shown to deteriorate under imbalanced data \cite{hevapathige2021evaluation,velayuthan2023revisiting}.

 Although various solutions have been proposed to mitigate class imbalance, they often suffer from issues such as being decoupled from model training or causing overfitting \cite{kishanthan2025deep}, making them less reliable under deep learning training conditions and less effective in real-world applications. Thus, we look for alternatives to tackle class imbalance from a novel perspective.  We observe that activation functions in deep learning possess strong capacity to introduce learning priors. However, existing traditional activation functions are primarily confined to introducing non-linearity into the model. To go beyond this, we propose incorporating learning priors through the activation function to help the model mitigate class imbalance. Our solution is built upon two key properties.

\paragraph{Orthogonality} Orthogonality, the property of feature independence, enhances the distinguishability of feature vectors in the embedding space, improving gradient flow and making model training more efficient \cite{kamalov2021orthogonal,karunasingha2023oc}. In the context of class imbalance, orthogonal feature spaces allow each class, including minority classes, to maintain its distinctiveness by occupying its own dimensions in the feature space, without being impacted from domination by the majority classes. Furthermore, orthogonal transformations preserve information during transformations, which is especially crucial for minority class instances, where each training example carries significant weight.

\paragraph{Group-Aware Bias} In the presence of class imbalance, maintaining proper distances between classes in the embedding space is crucial to prevent minority class instances from being overshadowed by the majority class. Ensuring that samples within the same class are closer together while maintaining adequate separation between different classes leads to a well-structured clustering in the embedding space. This clustering helps the classifier distinguish classes more effectively, without being biased by majority class instances. Learning a bias for each sample that aligns with this concept can shift embeddings, improving class separability and enhancing performance, while adapting to the varying density and distribution characteristics of different classes, offering more flexibility than a single global bias.

In light of this, we introduce a novel activation function, \emph{OGAB}, that incorporates Orthogonality and Group-Aware Bias into deep learning model training. This function can be seamlessly integrated into any model after each neural layer. Unlike existing solutions for class imbalance, our activation function does not require label information and is implicitly optimized through the downstream task loss. This eliminates the need for additional supervision or manual tuning, offering greater flexibility and scalability. Furthermore, by integrating directly into the model architecture, it enhances performance without requiring separate training steps or extensive pre-processing, making it more generalizable. We evaluate our solution on both real-world and synthetic imbalanced datasets, demonstrating its effectiveness against existing activation functions.

\section{Related Work}

\subsection{Techniques for Addressing Class Imbalance} Various approaches have been suggested to tackle the class imbalance, primarily through data resampling \cite{khushi2021comparative} and cost-sensitive learning \cite{ling2008cost}. Data resampling aims to equalize class distributions by employing minority oversampling \cite{kovacs2019empirical} and/or majority undersampling \cite{yen2006under}. In contrast, cost-sensitive learning introduces penalty mechanisms that encourage the model to focus more on minority class instances, improving predictive performance on imbalanced datasets \cite{khan2017cost}. Nevertheless, several challenges persist. Data resampling techniques, often used as a preprocessing step, are decoupled from the model training process, making them less adaptive to the learning procedure. Moreover, there is no guarantee that the resampled data will retain the semantic alignment of the original data. On the other hand, cost-sensitive learning introduces optimization challenges, including overfitting, reduced generalizability, and training instabilities. Prominently employed cost-sensitive learning methods include loss functions such as focal loss \cite{lin2017focal}, IoU loss \cite{yu2016unitbox}, GIoU loss \cite{rezatofighi2019generalized}, and orthogonality constraints \cite{karunasingha2023oc}. However, these approaches are prone to overfitting as they require careful tuning of parameters.  Further, the reweighting process in these cost-sensitive approaches could overweight minority classes, leading to amplified noise and outliers.

Unlike existing class imbalance techniques, our method does not explicitly rely on class labels or prior data distribution, making it broadly applicable and robust to label noise.

\subsection{Activation Functions}

Activation functions are mathematical functions which is applied to neural network layer outputs. In deep learning, activation functions are prominently employed to
introduce non-linearity to neural network layer outputs, allowing them to adapt complex patterns, thereby enhancing model capacity \cite{sharma2017activation}. Activation functions commonly employed in neural networks encompass a variety of options, including but not limited to, Rectified Linear Unit (ReLU), Sigmoid, Hyperbolic Tangent (Tanh), Softmax, and Softplus\cite{dubey2022activation}. While traditional activation functions are non-trainable, recent research has explored activation functions with trainable parameters. These trainable parameters provide adaptability during model training, enabling the functions to better capture complex patterns and leading to more generalizable models \cite{apicella2021survey,dubey2022activation}. Nonetheless, these trainable activation functions face challenges such as limited interpretability and potential dataset-specific performance, which can restrict their broader applicability.

Our method extends traditional activation functions with learnable parameters that are jointly optimized with the task loss, enhancing adaptability in deep learning. Unlike existing learnable activations, we leverage orthogonal feature separation and implicit group-aware bias adjustment to address class imbalance in a more effective and targeted manner.

\begin{figure*}[ht]
    \centering

\captionsetup{
justification=centering,
font=footnotesize, 
labelfont=rm
}

\includegraphics[width=0.9\textwidth, height=0.36\textheight]{Figures/OGAB_Pipeline_Updated.pdf} \caption{OGAB Architecture: (1) \textbf{Orthogonal Transformation:} The input feature undergoes an orthogonal transformation that preserves geometric relationships. (2) \textbf{Group-Aware Bias Learning:} A gating function assigns each input sample to a group, which then receives a customized bias. (3) \textbf{Combine Step:} The orthogonal transformation and biases are combined to produce the final output. 
}
    \label{fig:pipeline}
\end{figure*}

\section{Methodology}

Our activation layer, OGAB, consists of two phases: \emph{Orthogonal Transformation} and \emph{Implicit Group-Aware Bias Learning}.

\subsection{Orthogonal Transformation}

Let \( X \in \mathbb{R}^{n \times d} \) represent the input data matrix, where \( n \) is the batch size and \( d \) is the input dimensionality. We define several learnable parameters: a matrix \( A \in \mathbb{R}^{d \times d} \), and a scaling factor vector \( s \in \mathbb{R}^{d} \).

We construct an orthogonal matrix \( Q \) from the skew-symmetric matrix \( S \), which is derived from \( A \) as follows:
\begin{equation}
   S = A - A^T  
\end{equation}
Next, we apply the Cayley transform \cite{trockmanorthogonalizing} to obtain the orthogonal matrix \( Q \):
\begin{equation}
 Q = (I + S)(I - S)^{-1} 
\end{equation}
where \( I \in \mathbb{R}^{d \times d} \) is the identity matrix. \( Q \) satisfies the orthogonality condition \( Q^T Q = Q Q^T = I \), preserving vector norms (i.e. $| X Q |_2 = | X |_2$). This condition ensures stable gradient flow, prevents issues like exploding or vanishing gradients, and aids in efficient learning and faster convergence. Importantly, this is especially beneficial for data with class imbalance, as it ensures that the minority class retains its unique information during transformations, preventing any loss of significance due to the transformation.

To derive the orthogonal transformation feature matrix, we apply the orthogonal matrix 
$Q$ to the input, which can be expressed as:

\begin{equation}
U = X Q^T
\end{equation}

\subsection{ Implicit
Group-Aware Bias Learning}

We have the same \( X \in \mathbb{R}^{n \times d} \) as the input matrix. we define \( G \) groups, each associated with a bias vector of size \( d \). The group-specific bias vectors are stored in the matrix \( F \in \mathbb{R}^{G \times d} \), where the \( g \)-th row represents the bias for group \( g \). Additionally, the we define a learnable weight matrix \( M \in \mathbb{R}^{d \times G} \) and a bias vector \( B \in \mathbb{R}^G \), which together form a gating function that computes the probability distribution over groups for each sample.

For each sample \( X_i \) where \( i \in [1, n] \), we compute the probability \( p(i,g) \) that the sample belongs to group \( g \in G \) (i.e. gate probability), as follows:
\begin{equation}
\begin{aligned}
g_i &= M^T X_i + B \\
p(i, g) &= \frac{\exp(g_i)}{\sum_{k=1}^{G} \exp(g_k)}
\end{aligned}
\end{equation}

Each group has a bias vector \( F_g \in \mathbb{R}^{d} \). To assign a specific bias to each sample based on its group, we first expand the group biases across the batch, creating a tensor \( E \in \mathbb{R}^{n \times G \times d} \). Then, we weight the bias vectors by the corresponding group probabilities \( p(i) = (p(i,1), \dots, p(i,G)) \) for each sample. This is done by element-wise multiplying the gate probabilities \( p(i) \) with the expanded biases. To ensure proper broadcasting, \( p(i) \in \mathbb{R}^{G \times 1} \) is reshaped to \( \hat{p}(i) \in \mathbb{R}^{G \times d} \). The final bias for each sample \( V_i \in \mathbb{R}^{d} \) is then given by:

\begin{equation}
V_i = T\left( \hat{p}(i) \odot E_i \right)
\end{equation}

where \( \odot \) denotes element-wise multiplication, and \( T(\cdot) \) represents the summation across all groups to produce the final  bias.

\subsection{Combination Step}

We combine the orthogonal transformation and the group-ware bias to derive the final output $Y$ as follows:
\begin{equation}
W = U + V
\end{equation}
\begin{equation}
Y = s \odot \sigma(W)
\end{equation}
where \( \odot \) denotes element-wise multiplication. The transformation \( \sigma(\cdot) \) complements the model by adding additional non-linearity, while the element-wise scaling \( \mathbf{s} \) allows flexible output adjustment. Additionally, it is desirable for \( \sigma(\cdot) \) to be differentiable and smooth for stable gradient propagation during training.

\subsection{Runtime Complexity Analysis}

OGAB has a runtime complexity of $\mathcal{O}(d^3 + nd^2 + nGd)$ per forward pass. The computation of the orthogonal matrix requires $\mathcal{O}(d^3)$ operations, including matrix inversion, while applying the transformation costs $\mathcal{O}(nd^2)$. Additionally, group-aware bias learning adds $\mathcal{O}(nGd)$ for gate computation and bias assignment.

The high-level architecture of OGAB is depicted in Fig \ref{fig:pipeline}.

\section{Experiments}

\subsection{Datasets}

To evaluate the performance of OGAB, we utilize six datasets: five real-world datasets \cite{karunasingha2023oc, menzies2004assessing, straw2022investigating} and one synthetic dataset \cite{guyon2003design}. A summary of these datasets is provided in Table \ref{tab:dataset_stats}. IR denotes the imbalance ratio.
\begin{table}[h!]
\captionsetup{
  justification=centering,
  font=footnotesize,
  labelsep=newline,
  singlelinecheck=false
}
\centering
\caption{SUMMARY OF DATA SETS}
\begin{tabular}{|l|c|c|c|c|}
\hline
\textbf{Data Set} & \textbf{Type} & \textbf{Size} & \textbf{\# Features} & \textbf{IR} \\ \hline
Thyroid  & Binary & 7,200 & 21 & 12.5 \\ \hline
KC1  & Binary & 2,109 & 21 & 5.5 \\ \hline
ILPD  & Binary & 583 & 10 & 2.5 \\ \hline
Page Blocks  & Multi-Class & 5,473 & 10 & 175.5 \\ \hline
Pen Digits  & Multi-Class & 1,100 & 16 & 2.2 \\ \hline
Synthetic & Multi-Class & 10,000 & 20 & 16.0 \\ \hline
\end{tabular}
\label{tab:dataset_stats}
\end{table}

\subsection{Classifier}

We employ a Multi-layer Perceptron (MLP) \cite{ramchoun2016multilayer}  as our deep learning classifier. Further, we use cross-entropy loss \cite{mao2023cross} as our training objective. The objective is trained to minimize cross-entropy $H(p, q)$ between the true label distribution (i.e. ground truth) \( p \) and the predicted label probability distribution  \( q \) is given by:

\begin{equation}
H(p, q) = - \sum_{i=1}^{N} p(c_i) \log(q(c_i))   
\end{equation}

where \( N \) is the number of possible classes, \( p(c_i) \) is the true probability distribution of the class \( c_i \), and \( q(c_i) \) is the predicted label probability distribution for class \( c_i \).

\subsection{Experimental Setup and Baselines}

 The training data is split with an 80:20 ratio for training and testing. Features in each dataset are normalised using min-max scale \cite{priyambudi2024algorithm} in a pre-processing step. The model hyperparameters are as follows: a learning rate of 0.01, 500 training epochs, a batch size of 500, 4 deep learning layers, 64 hidden dimension, no dropout, and \( G \in \{1, 5, 10\} \). We use the Adam optimizer \cite{kingma2014adam} to efficiently update the model's parameters during training. Each deep learning model is trained 10 times with different random seed values, and the average of each evaluation metric is reported.

For the baselines, we employ commonly used traditional non-learnable activation functions, namely, ReLU, Softmax, Tanh, Sigmoid, and SoftPlus, as well as learnable activation functions, including PReLU \cite{he2015delving}, ConvReLU \cite{gao2020adaptive}, and SLAF \cite{goyal2019learning}.

\begin{table*}[h!]
\captionsetup{
  justification=centering,
  font=footnotesize,
  labelsep=newline,
  singlelinecheck=false
}
\centering
\caption{PERFORMANCE COMPARISON OF ACTIVATION FUNCTIONS ACROSS DIFFERENT DATASETS. THE BEST RESULTS ARE HIGHLIGHTED IN \textbf{BOLD}. NOTE THAT THE BASELINE REFERS TO THE DEEP LEARNING MODEL WITHOUT ANY ACTIVATION FUNCTION.}

\renewcommand{\arraystretch}{1.0}
\begin{tabular}{|l|c|c|c|c|c|c|c|c|c|c|c|c|}
\hline
 & \multicolumn{2}{c|}{Thyroid} & \multicolumn{2}{c|}{KC1} & \multicolumn{2}{c|}{ILPD} & \multicolumn{2}{c|}{Page-blocks} & \multicolumn{2}{c|}{Pen Digits} & \multicolumn{2}{c|}{Synthetic} \\ \hline
 & F1-score & B. Acc. & F1-score & B. Acc. & F1-score & B. Acc. & F1-score & B. Acc. & F1-score & B. Acc. & F1-score & B. Acc. \\ \hline
Baseline & 83.79 &  78.53 &  62.15 & 59.67 & 56.19 &  56.69 & 80.46 & 76.35 & 94.33 & 94.35 & 57.87 & 53.02 \\ 
\hline
 \multicolumn{13}{c}{Non-learnable Activation Functions} \\
 \hline
ReLU & 94.71 & 93.79 & 65.23 & 63.21 & 58.85 & 59.22 &  82.37  & 81.13 &  99.30 & 99.30 & 94.71 & 94.01\\ \hline
Tanh &  85.84 & 86.47 &   64.03 & 61.91 &  54.23 & 56.07 & 82.52 & 81.57 &  98.34 & 99.24 & 94.10 & 92.73\\ \hline
Softmax & 48.07 & 50.00 &   54.81 & 55.42 & 41.50 & 50.00 & 75.19 & 73.65 &   94.79 & 94.83 & 87.13 & 86.52\\ \hline
Sigmoid & 95.20 & 95.38 &  62.18 & 59.75 & \textbf{61.20} & \textbf{60.61} & 83.80 & 81.70 & 99.24 & 99.23 & 90.84 & 89.91 \\ \hline
SoftPlus & 94.88 & 95.71 & 62.41 & 59.88 & 58.80 & 58.67 & 82.64 & 82.80 & 98.41 & 99.31 & 94.15 & 93.33 \\ \hline
 \multicolumn{13}{c}{Learnable Activation Functions} \\
 \hline
 PReLU & 94.12 & 93.09 & 66.34 & 64.21 & 59.48 & 59.48 & 83.78 & 83.72 & 99.38 & 99.38  & 94.61 & 93.53 \\ \hline
  ConvReLU & 95.37 & 96.57 & 66.80 & \textbf{65.03} & 58.93 & 58.99 & 83.14 &  83.31 & 99.40 & 99.40 & 94.99 & 94.01 \\ \hline
  SLAF & 72.72 & 73.86  & 61.97 & 59.61 & 59.29 & 59.15 & 81.41 & 80.20 & 99.37 & 99.37 & 93.54 &  92.14 \\ \hline
OGAB & \textbf{95.46} & \textbf{96.67} & \textbf{66.80} & 64.83 & 59.76 & 59.57 &  \textbf{84.73}  & \textbf{84.95} & \textbf{99.43} & \textbf{99.43} & \textbf{96.59} & \textbf{95.56} \\ \hline
\end{tabular}
\label{tab:classification}
\end{table*}

\subsection{Evaluation Metrics}

We use F1-score and balanced accuracy as evaluation metrics due to their sensitivity to class imbalance \cite{diallo2024machine} . The number of True Positives (TP), True Negatives (TN), False Positives (FP), and False Negatives (FN) predicted by the classifier are used to compute these metrics, with the corresponding equations stated as follows.

\begin{equation}
\text{F1-score} = 2 \times \frac{TP}{2TP + FP + FN}   
\end{equation}

\begin{equation}
\text{Balanced Accuracy} = \frac{1}{2} \left( \frac{TP}{TP + FN} + \frac{TN}{TN + FP} \right)
\end{equation}

\subsection{System Resources and Implementation Details}

We employ Python programming language for implementations, along with the Scikit-learn and PyTorch libraries \cite{pajankar2022hands}. All experiments were conducted on a MacBook equipped with an Apple M2 chip, 8 GB of unified memory, and a 10-core GPU. 

\section{Results and Discussion}

\subsection{Classification Performance Evaluation}

We present the classification results for the baselines and OGAB in Table \ref{tab:classification}. Traditional activation functions exhibit varying performance across different datasets: Sigmoid performs well on the Thyroid dataset, ReLU excels with the Pen Digits dataset, and Tanh shows strong results on the Page-Blocks dataset. In contrast, OGAB demonstrates superior performance across all datasets, surpassing or providing competitive performance with traditional activation functions. When compared to other learnable activation functions, OGAB still maintains its edge. While PReLU, ConvReLU, and SLAF show improvements over the baseline and some traditional activations, OGAB consistently outperforms them across metrics. Notably, on the Synthetic dataset, OGAB achieves the highest F1-score (96.59\%) and balanced accuracy (95.56\%), significantly outperforming both ConvReLU and PReLU.  

The enhanced performance of OGAB can be attributed to its learnability and adaptive design. Unlike fixed activation functions and even compared to other learnable alternatives, OGAB adapts its behavior more effectively to the unique characteristics of each dataset. This superior adaptability enables it to optimize performance regardless of the underlying data distribution.

\subsection{Ablation Study}

We evaluate the contribution of the two main components in OGAB to its performance by removing each component and measuring the classification performance. The results for a subset of datasets are presented in Table \ref{tab:ablation_study}, reporting the F1-score. It is evident that each component contributes uniquely to the performance, as the best results are consistently achieved by the model with both components. Additionally, it can be observed that the orthogonality component contributes slightly more compared to the group-aware bias, as seen from the comparatively large performance drop when orthogonality is removed.  

\begin{table}[H]
\captionsetup{
  justification=centering,
  font=footnotesize,
  labelsep=newline,
  singlelinecheck=false
}
\centering
\caption{PERFORMANCE COMPARISON OF DIFFERENT COMPONENTS IN OGAB. THE BEST RESULTS ARE HIGHLIGHTED IN \textbf{BOLD}}

\begin{tabular}{|l|c|c|c|}
\hline
 & Thyroid & ILPD & Page Blocks \\
 \hline
ReLU & 94.71 & 58.85 &  82.36 \\
\hline
OGAB w/o orthogonality & 94.88 & 58.58 & 82.39\\
\hline
OGAB w/o group-aware bias & 95.06 & 58.74 & 83.26\\
\hline
OGAB & \textbf{95.46} & \textbf{59.76} & \textbf{84.72}\\
\hline
\end{tabular}

\label{tab:ablation_study}
\end{table}

\subsection{Comparison with Sampling Approaches}

We compare the performance of OGAB with existing sampling approaches commonly used to address the class imbalance problem. Specifically, we evaluate our method against SMOTE \cite{chawla2002smote}, Tomek Links \cite{pereira2020mltl}, and SMOTEENN \cite{puri2022improved}, which represent widely used oversampling, undersampling, and hybrid sampling techniques, respectively.
\subsubsection{Performance Comparison} 
\begin{figure}[ht]
    \centering
     \captionsetup{justification=centering,
     font=footnotesize, 
    labelfont=rm
    }
    \includegraphics[width=0.8\columnwidth]{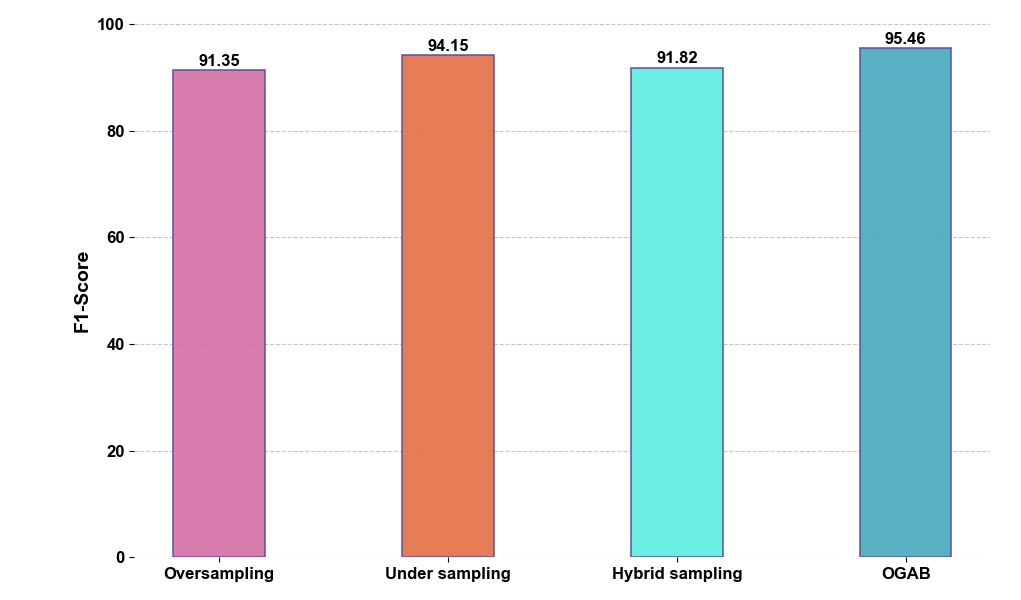} 
    \caption{Performance Comparison with Sampling Approaches}  
    \label{fig:F1_Score_Comparison}
    \vspace{-10pt} 
\end{figure}

The classification performance comparison for Thyroids dataset is depicted in Fig \ref{fig:F1_Score_Comparison}. OGAB outperforms other sampling approaches mainly due to its coupled behaviour with the learning task, enabling more effective model training.

\subsubsection{Runtime Complexity Comparison} 
\begin{figure}[ht]
    \centering
     \captionsetup{justification=centering,
     font=footnotesize, 
labelfont=rm
     }
     
    \includegraphics[width=0.8\columnwidth]{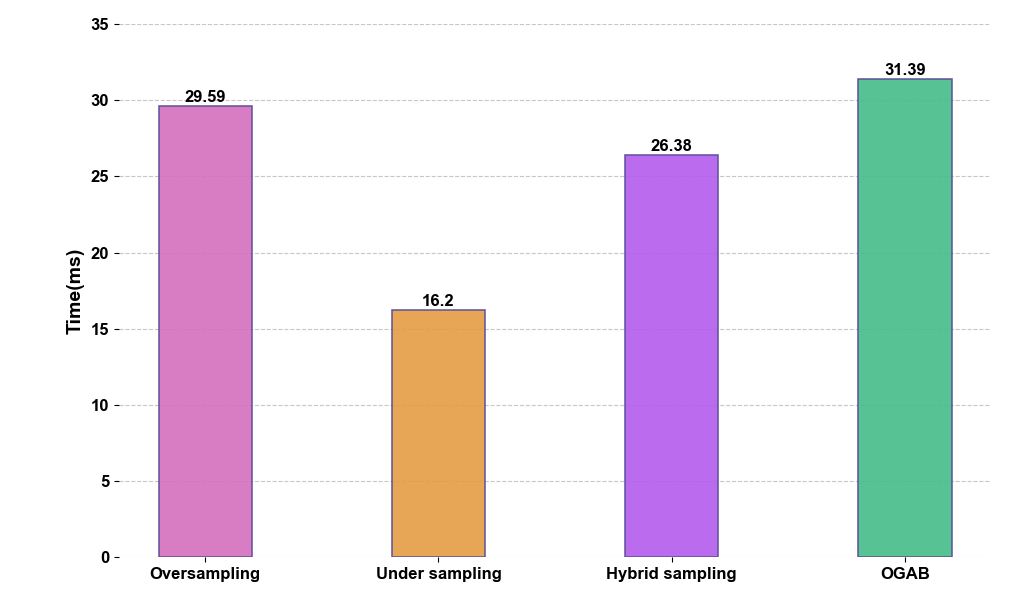} 
    \caption{Runtime Comparison with Sampling Approaches}  
    \label{fig:Time_Comparison}
    \vspace{-10pt} 
\end{figure}

We compare the runtime of OGAB with other sampling approaches by reporting the model training time per epoch on the Thyroid dataset, as shown in Fig \ref{fig:Time_Comparison}. The results indicate that the runtime of OGAB is comparable to that of other sampling methods, such as SMOTE and SMOTEENN.

In a nutshell, OGAB offers an effective solution to the class imbalance problem compared to existing methods, without compromising computational efficiency.

\subsection{Parameter Complexity Analysis}

\begin{figure}[ht]
    \centering
     \captionsetup{justification=centering,
     font=footnotesize, 
labelfont=rm}
    \includegraphics[width=0.8\columnwidth]{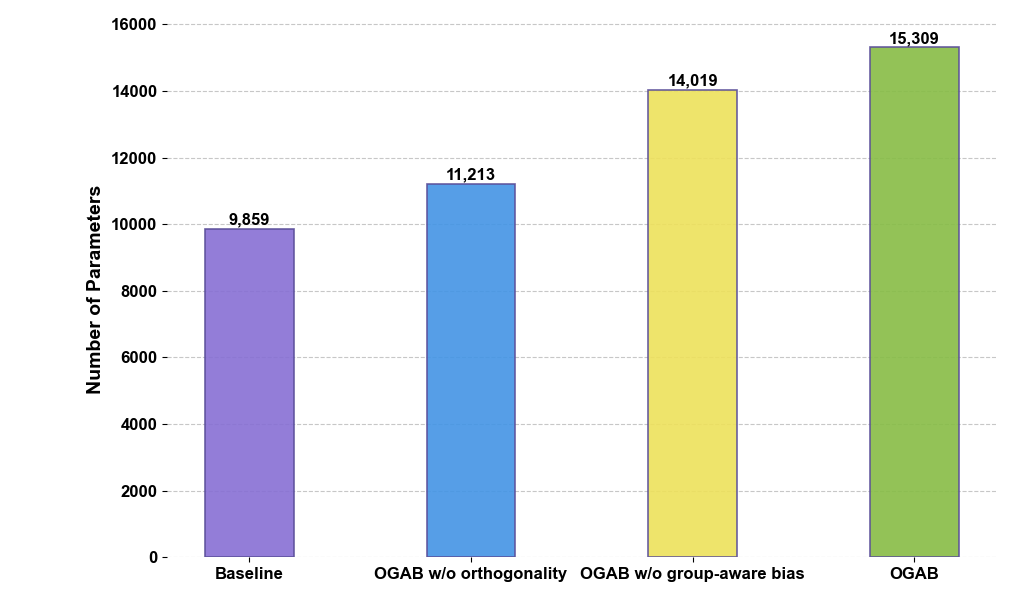} 
    \caption{Parameter Count Comparison for the Baseline and Different OGAB Variants}  
    \label{fig:Parameter_Complexity_Analysis}
    \vspace{-10pt} 
\end{figure}
We compare the number of learnable parameters in deep learning models with OGAB variants to those in models with traditional non-learnable activation functions, as shown in Fig \ref{fig:Parameter_Complexity_Analysis}. OGAB incurs an approximately 55\% increase in the number of learnable parameters. Among its two components, the orthogonal component has a larger parameter count than the group-aware bias component. This is justified by the ablation study, which demonstrates that the orthogonal component contributes more significantly to the performance of OGAB.

\subsection{Visualization Analysis}
\begin{figure}[ht]
    \centering
     \captionsetup{justification=centering,
     font=footnotesize, 
    labelfont=rm}
    \begin{subfigure}[b]{0.48\columnwidth}
        \centering
        \includegraphics[width=\linewidth]{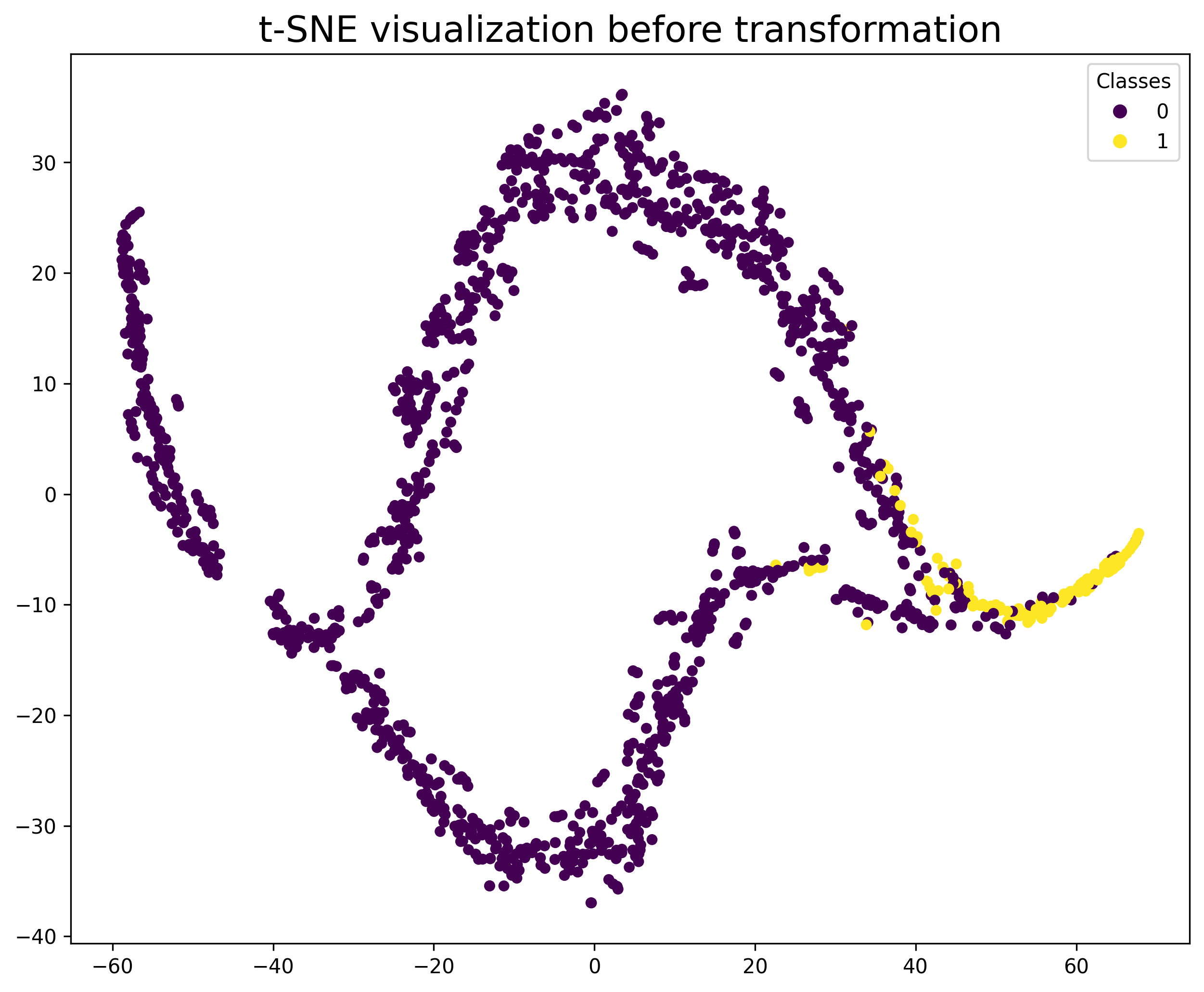}
        \caption{Binary classification - before transformation}
        \label{fig:Thyroid_before_tranf}
    \end{subfigure}
    \hfill
    \begin{subfigure}[b]{0.48\columnwidth}
        \centering
        \includegraphics[width=\linewidth]{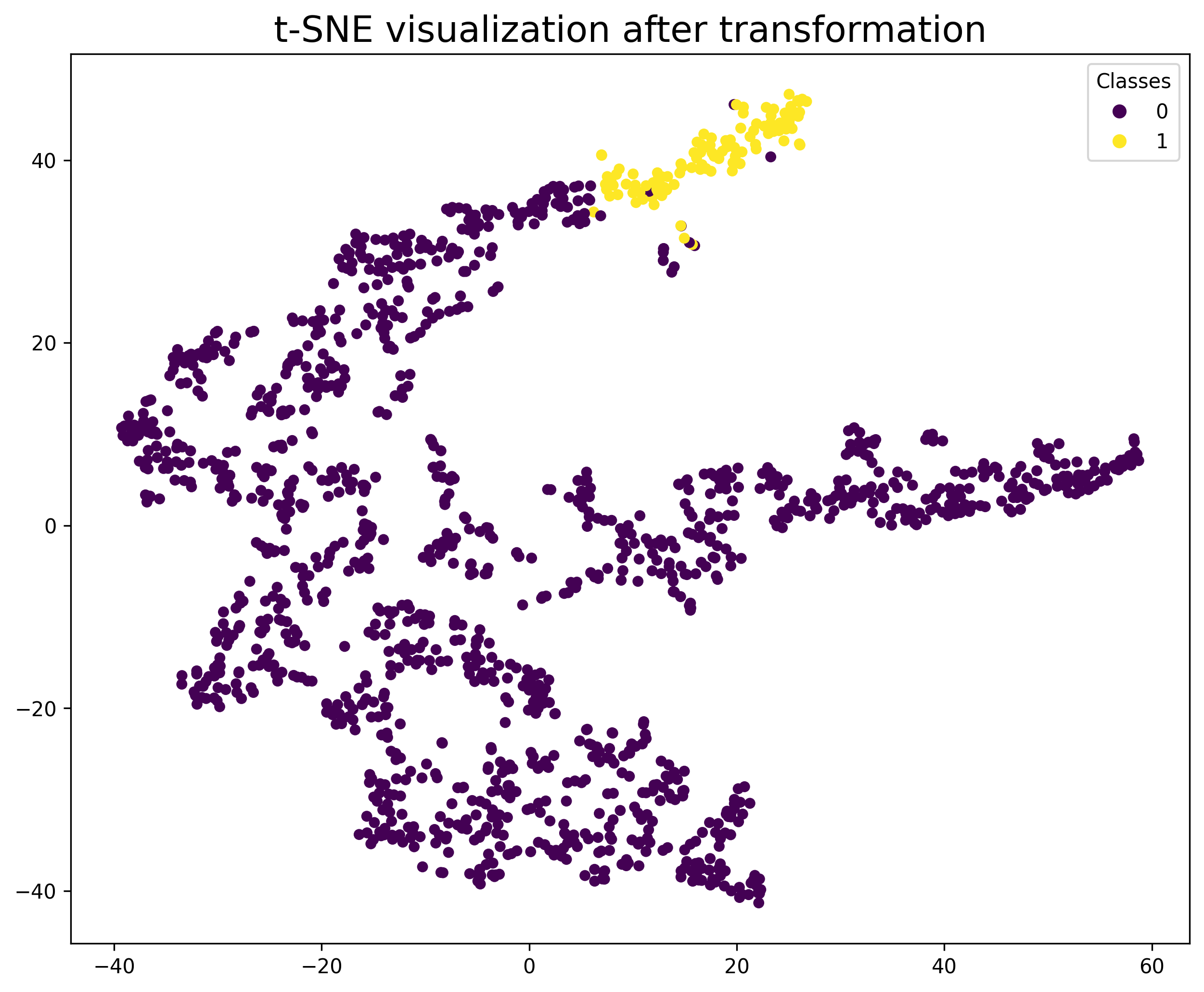}
        \caption{Binary classification - after transformation}
        \label{fig:Thyroid_after_tranf}
    \end{subfigure}
    
    \vspace{0.5cm}
    
    \begin{subfigure}[b]{0.48\columnwidth}
        \centering
        \includegraphics[width=\linewidth]{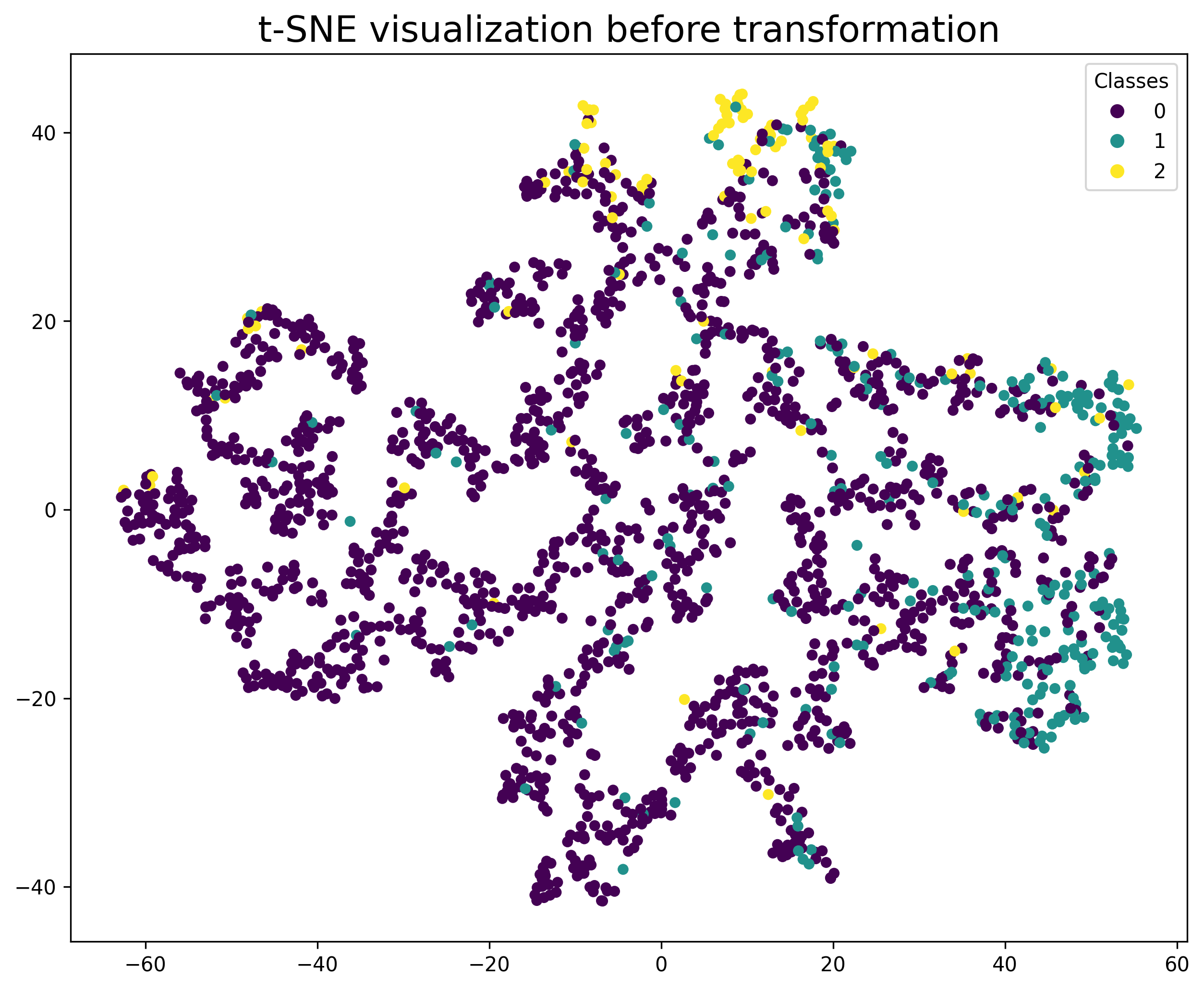}
        \caption{Multi class classification - before transformation}
        \label{fig:Synthetic_before_tranf}
    \end{subfigure}
    \hfill
    \begin{subfigure}[b]{0.48\columnwidth}
        \centering
        \includegraphics[width=\linewidth]{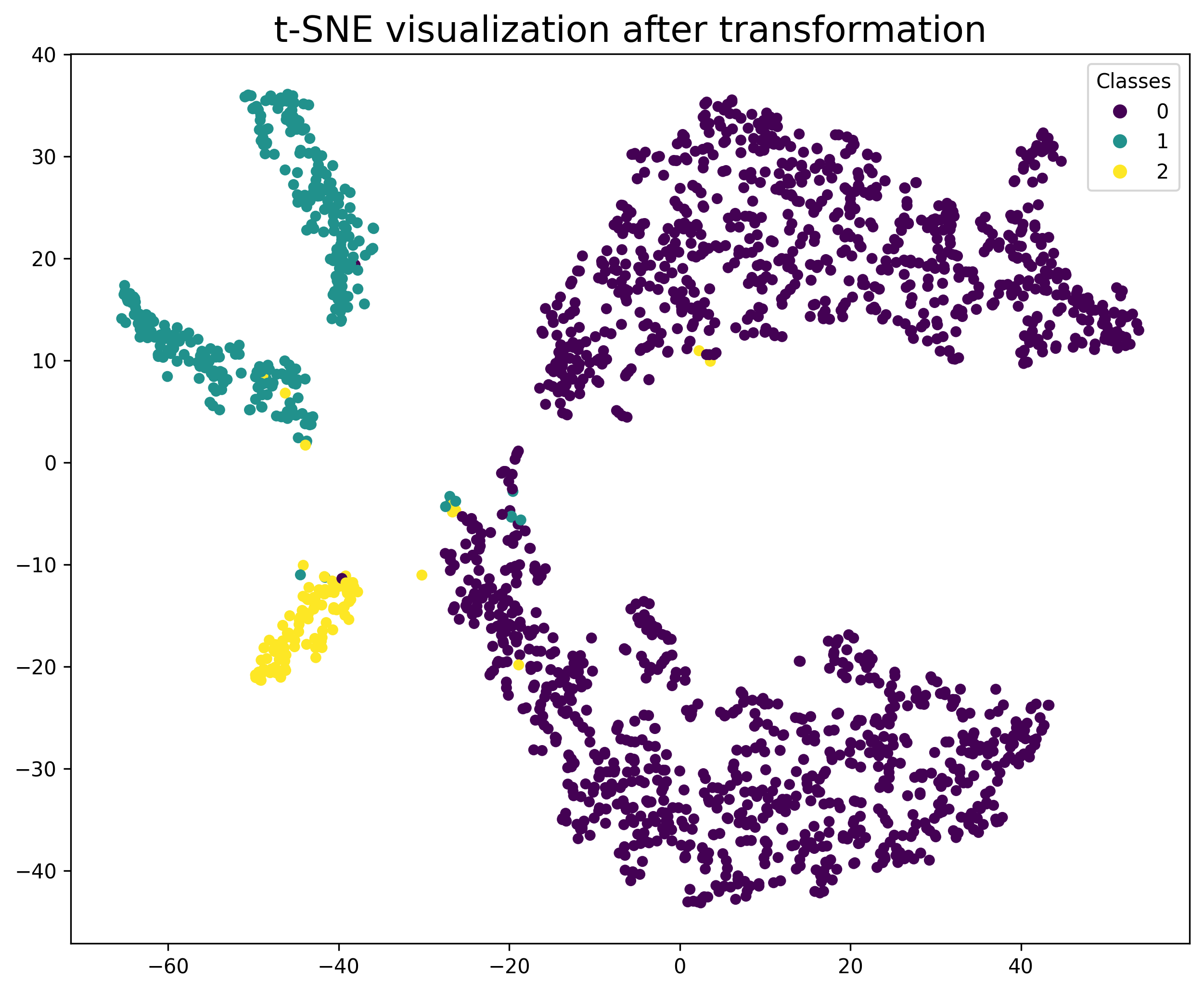}
        \caption{Multi class classification - after transformation}
        \label{fig:Synthetic_after_tranf}
    \end{subfigure}
    \caption{Embedding Visualization}
    \label{fig:Parameter_Complexity_Analysis}
    \vspace{-10pt}
\end{figure}
In Fig \ref{fig:Parameter_Complexity_Analysis}, we analyze the impact on the embedding after applying OGAB using t-SNE visualization \cite{arora2018analysis}. We visualize the 2D embedding space for the Thyroid and Synthetic datasets, representing binary and multi-class classification scenarios, respectively. In both instances, OGAB provides improved class separation. The orthogonality ensures that embeddings in different classes are perpendicular and independent, while the group-aware bias brings embeddings of the same class closer together and maintains proper distance between embeddings of different classes. This creates a well-defined clustering structure in the embedding space, ultimately enhancing classifier prediction performance.

\section{Conclusion, and Future Work}

In this work, we introduce a novel activation function designed to address class imbalance in deep learning models. The proposed solution leverages two key properties: feature orthogonality and feature group separability, defined via a bias term. We demonstrate that our solution outperforms traditional activation functions on imbalanced classification task. 

In our future work, we plan to combine our solution with transfer learning to achieve better generalization in real-world data. Additionally, we plan to implement more computationally efficient versions of OGAB with reduced parameter budgets.

\bibliography{references} 
\bibliographystyle{ieeetr}

\end{document}